%% file: main.tex
\documentclass[10pt,twocolumn,letterpaper]{article}

\usepackage{iccv}              % To produce the CAMERA-READY version

\definecolor{iccvblue}{rgb}{0.21,0.49,0.74}
\usepackage[pagebackref,breaklinks,colorlinks,allcolors=iccvblue]{hyperref}
\def\paperID{5} % *** Enter the Paper ID here
\def\confName{ICCV}
\def\confYear{2025}

\title{Locally Controlled Face Aging with Latent
Diffusion Models}

\author{
Lais Isabelle Alves dos Santos \and
Julien Despois \and
Thibaut Chauffier \and
Sileye O. Ba \and
Giovanni Palma\\[0.5em]
L'Oréal AI Research\\
}

\begin{document}
\maketitle

\begin{figure*}[!htb]
\centering
\includegraphics[width=0.85\linewidth]{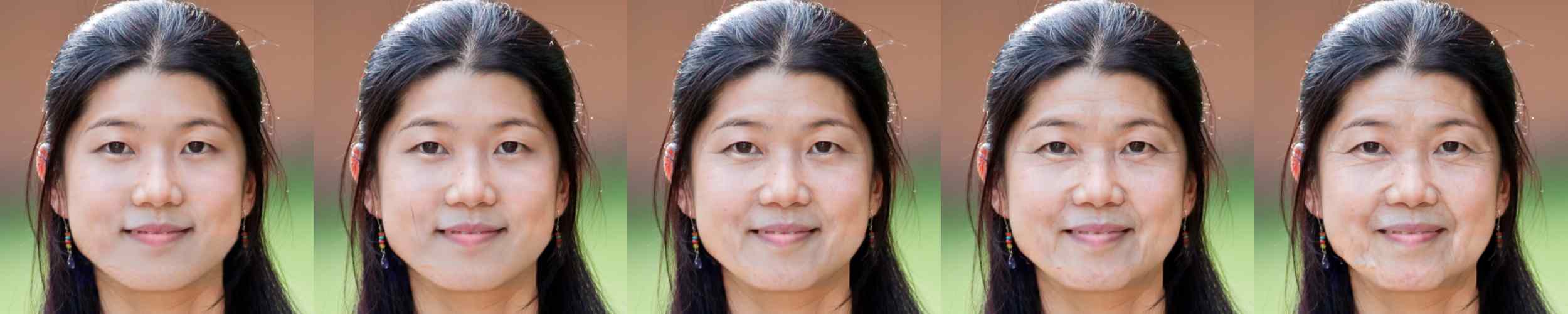}
\captionsetup{justification=centering, margin=0.7cm}
\caption{Controlled face wrinkles evolution with the proposed method}
\label{fig:intro-image}
\end{figure*}

\input{sections/0_abstract}
\input{sections/1_intro}
\input{sections/2_literature}
\input{sections/3_model}
\input{sections/4_datasets}
\input{sections/5_results}

\input{sections/6_conclusion}
% \input{sections/X_suppl}

{
    \small
    \bibliographystyle{ieeenat_fullname}
    \bibliography{main}
}

% \input{sections/X_suppl}

\end{document}

%% file: sections/0_abstract.tex
\begin{abstract}
% The ABSTRACT is to be in fully justified italicized text, at the top of the left-hand column, below the author and affiliation information.
% Use the word ``Abstract'' as the title, in 12-point Times, boldface type, centered relative to the column, initially capitalized.
% The abstract is to be in 10-point, single-spaced type.
% Leave two blank lines after the Abstract, then begin the main text.
% Look at previous \confName abstracts to get a feel for style and length.
We present a novel approach to face aging that addresses the limitations of current methods which treat aging as a global, homogeneous process. Existing techniques using GANs and diffusion models often condition generation on a reference image and target age, neglecting that facial regions age heterogeneously due to both intrinsic chronological factors and extrinsic elements like sun exposure. Our method leverages latent diffusion models to selectively age specific facial regions using local aging signs. This approach provides significantly finer-grained control over the generation process, enabling more realistic and personalized aging. We employ a latent diffusion refiner to seamlessly blend these locally aged regions, ensuring a globally consistent and natural-looking synthesis. Experimental results demonstrate that our method effectively achieves three key criteria for successful face aging: robust identity preservation, high-fidelity and realistic imagery, and a natural, controllable aging progression.
\end{abstract}

%% file: sections/1_intro.tex
\section{Introduction}
Face aging is an image generation task that aims to transform a reference image of a person into a realistic, aged version of the same individual. Well executed, it enables numerous real-world applications in augmented reality, cinema, and the cosmetic industry.
Research on face aging generation typically focuses on controlling the generated output by using a person's age as input. However, it should be noted that aging is complex and manifests in different ways across different populations and individuals. It varies across different parts of the face \cite{atlas1,atlas2, atlas3, atlas4, atlas5}, and is influenced by factors and lifestyle choices, such as smoking, sun exposure, and alcohol consumption \cite{defining-skin-aging, raynaud2024quantifying}. It is also influenced by daily facial expressions, which accelerate the appearance of aging signs, particularly in areas such as the forehead or around the mouth. 

Over the past few years, popular models such as Pix2Pix \cite{pix2pix} and StarGAN \cite{stargan1, stargan2} provided a way to translate images to other domains and condition the generation towards a desired output, including the task of face aging. However, these approaches lack control over the aging process and do not account for the fact that people age differently. AMGAN \cite{amgan} offers a way to control generation through more detailed aging maps but struggles to produce natural face wrinkles when the reference image lacks predefined aging marks.

More recently, diffusion models have demonstrated better performance than GANs in terms of image quality \cite{class-guide} with methods such as PADA \cite{pada} and FADING \cite{aging-editing} providing insights for face aging using diffusion models. However, they offer little to no control over the aging of different parts of the face.

To address these issues, we propose \textbf{L}atent \textbf{D}iffusion \textbf{L}ocal \textbf{A}ging (\textbf{LDLA}), a diffusion-based model that offers greater control over the zones of the face by using ethnicity-specific skin atlases \cite{atlas1, atlas2, atlas3, atlas4, atlas5}. These atlases define reference scales for the aging of every specific part of the face, such as crow's feet or forehead wrinkles. Our model performs face aging at a fine-grained level and leverages Latent Diffusion Models \cite{stable-diffusion} to transform each facial zone individually according to target scores. The model finally blends the individual zones and refines the output, producing high-quality images (Figure~\ref{fig:intro-image}). The key contributions of our work are as follows:

\begin{itemize}
    \item A novel use of latent diffusion models to achieve state-of-the-art fine-grained face aging.
    \item A latent cycle consistency loss using a single denoising step to preserve the subject's identity and improve efficiency.
    \item The use of a refiner model to blend the individual crops seamlessly, generating more realistic images.
\end{itemize}

%% file: sections/2_literature.tex
\section{Literature review}
\label{sec:literature-review}
Many methodological frameworks have been proposed for image generation, such as generative adversarial networks (GANs) \cite{goodfellow2014generative}, normalizing flows \cite{kingma2018glow}, and recently denoising diffusion probabilistic models \cite{diffusion-models}. All these models have conditional variants, where the generated images are based on specific input conditions. When the condition involves an image being transformed, the model is typically known as an image-to-image translation model.

\textbf{Generative Adversarial Networks} for image-to-image translation, such as CycleGAN \cite{Zhu2017cycle-gan}, introduced the cyclic generation methodology where, during training, the model generates the target from the source and then regenerates to the source from the target. This framework allows generative models to be trained without matching source-target pairs between two domains. StarGAN \cite{stargan1} further extended cyclical generation to any number of domains.
Despite their popularity, GAN usage is limited by the instability of their training due to the adversarial loss involving the competition between the generator and discriminator models. 

\textbf{Denoising Diffusion Probabilistic Models (DDPMs)} \cite{ddpm} have been introduced as an approach to simplify training. Unlike adversarial frameworks, these models use a forward and a backward process. In the forward process, Gaussian noise is gradually added to an image until total information destruction. In the backward process, a generative model learns to remove added noise to recover the original image. Guided and unguided diffusion models \cite{class-guide, class-free-guide} have been proposed to improve the diffusion process, by conditioning the generation process. 

Later, \textbf{Latent Diffusion Models (LDMs)} \cite{stable-diffusion} introduced the use of latent space embedding to improve training stability and computational efficiency. These models, known as Stable Diffusion, incorporate the diffusion and denoising processes within the latent space and condition the generation on both text and image inputs. In this paper, we adopt this architecture as the backbone of our model.

\textbf{Controllable Face aging} is a conditional image-to-image translation task where, given a person's face and age-related target information, a model generates an aged image of the person. Age has been explored as a key conditioning factor in various face-aging models. For instance, the StarGAN model \cite{stargan1, stargan2} incorporates age as one of its conditioning parameters. Similarly, \cite{preechakul2022diffusion-autoencoders} employs a latent diffusion model for face aging, where age acts as the conditioning input. In other approaches, such as \cite{Chen-Lathuiliere-2023face} and \cite{li2023-pluralistic-aging-ddpm}, age is encoded in a text prompt to guide the face aging process. However, face aging is a complex process, as different face zones (e.g., forehead, under-eye, mouth corners) may age at different rates, making it challenging to capture the phenomenon with age as a conditioning factor. One potential solution is to transform face zones individually using local skin aging labels \cite{atlas1, atlas2, atlas3, atlas4, atlas5} as proposed by AMGAN \cite{amgan}. This enables locally controlled face aging, allowing more personalized and precise generation. Therefore, we follow this fine-grained controlled approach over face aging generation to design our model leveraging latent diffusion models, and will use AMGAN as a baseline.

%% file: sections/3_model.tex
\section{Proposed model}
\label{sec:proposed-model}
We propose an approach based on latent diffusion models to achieve locally controlled face aging. Rather than relying on age as a parameter, the focus is placed on the natural evolution of face wrinkles by relying on locally normalized aging scores. Each part of the face is generated individually, using a target score as a condition for each zone, rather than generating the entire face as a whole. The generated crops are then blended to produce full-face images, an approach that yields smoother and more realistic results.

\subsection{Controlled face aging with Latent Diffusion Models}

\begin{figure*}[!h]
    \centering
    \includegraphics[width=1.0\linewidth]{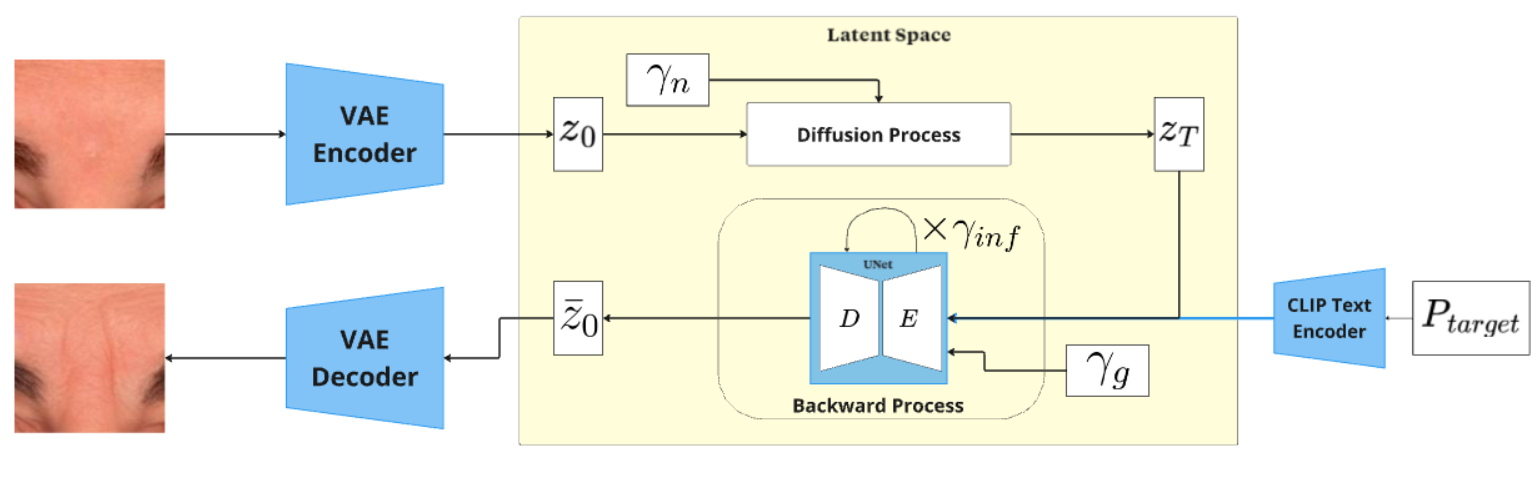}
    \captionsetup{justification=centering, margin=0.7cm}
    \caption{The schematic of the inference stage, regulated by three different hyperparameters.}
    \label{fig:model-inference-scheme}
\end{figure*}

Our model is based on the text-to-image LDMs, with several enhancements to achieve greater control over the aging of the generated images. Our model is divided into two stages, illustrated in Figures \ref{fig:model-train-scheme} and \ref{fig:model-inference-scheme}. The first stage involves fine-tuning the weights of the UNet Conditional model \cite{pre-trained-model-sd} by using crops from the different zones of the face, conditioned on target scores, and applying several losses to optimize the model toward producing natural, identity-preserving, controlled images. The second stage performs latent parameter-guided diffusion and denoising to generate the transformed image crop.

\subsubsection{Model training}

\begin{figure*}[!t]
    \centering
    \includegraphics[width=1.0\linewidth]{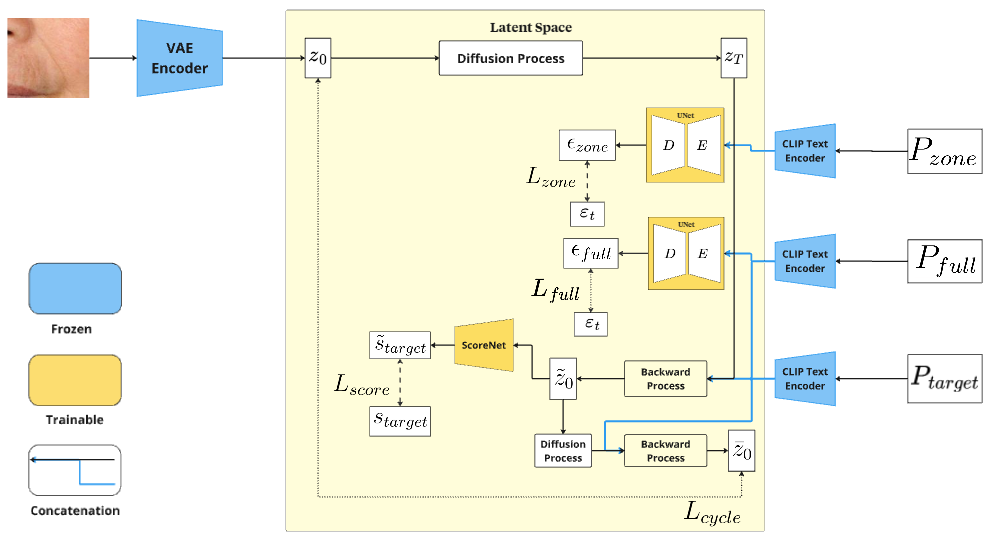}
    \captionsetup{justification=centering, margin=0.7cm}
    \caption{The UNet is optimized using $L_{full}$, $L_{zone}$, $L_{cycle}$, and $L_{score}$, while the ScoreNet is optimized uniquely with the score loss.}
    \label{fig:model-train-scheme}
\end{figure*}

In the first stage, we use a pre-trained stable diffusion model \cite{pre-trained-model-sd, HF-text2image} to leverage the learned information about facial geometry from the extensive LAION-5B dataset \cite{LAION-dataset}. The proposed architecture is illustrated in Figure~\ref{fig:model-train-scheme}. Our model retains the core structure proposed in Stable Diffusion \cite{stable-diffusion}, i.e., the denoising-diffusion occurs in the latent space, and the Variational Auto-Encoder (VAE) \cite{vae} and the CLIP Text Encoder \cite{clip} remain frozen. In this regard, our main contribution lies in the process by which the UNet is fine-tuned to generate high-quality, naturally aged or rejuvenated images.

% A DDIM Scheduler \cite{ddim} introduces noise to the input latent for a specific time step $t$, as part of the aforementioned diffusion process. The choice of this scheduler implies that the entire forward process will be much faster, as these implicit models rely on a non-Markovian chain, which allows access to multiple past-time steps.

Cross-attention in Stable Diffusion's UNet leads to better spatial awareness of the tokenized conditioning prompt \cite{stable-diffusion}. Consequently, our model employs three distinct prompts. The first prompt, $P_{full}$, ensures the model performs the task from a global perspective. The second prompt, $P_{zone}$, helps the model identify the specific crop zone and focus on increasing or decreasing wrinkles. Finally, the third prompt, $P_{target}$, is used to guide the initial step of the latent cycle consistency loss.

The primary prompt, $P_{full}$, contains all the necessary information to train the model. This prompt provides details about the input image $x_0$, such as the zone of the face, the person’s self-reported ethnicity, and the aging score of the current crop, expressed as a percentage. This percentage is obtained by dividing the local aging score by the maximum of the scale as defined in the atlases. An example would be  ``\textit{image of forehead wrinkles with an aging score of 70\% for a person of Hispanic ethnicity}''. By employing percentages, we leverage the pre-trained CLIP encoder's learned semantics instead of having to provide the lower and upper bounds for the scales as defined in the skin atlases, which the model would have to learn. The target level of noise $\epsilon_t$ for a time step $t$ is then predicted as $\epsilon_{full}$ by conditioning the UNet on $\tau_\theta(P_{full})$, $\tau_\theta$ being the text encoder, and $z_t$ the noisy latent embedding of the input image. With that, a Mean Square Error (MSE) Loss is computed over the noises and defined in Equation~ (\ref{eq_loss-full}). In summary, the goal of the training phase is to learn what each of the features present in the prompt represents in the latent space and how to reproduce them later in the inference phase.

\begin{equation} \label{eq_loss-full}
    L_{full} = \mathbb{E}_{t, P_{full}, \epsilon_t \thicksim \mathcal{N}(0, \mathbf{I})} 
        \Big [ \| \epsilon_t - \epsilon_\theta (z_t, t, \tau_\theta(P_{full})) \|_2^2 \Big]
\end{equation}

The second prompt, $P_{zone}$, guides the model to learn the distinctive features of each facial zone, as crops from different zones may appear similar but age in distinct ways. For example, a smooth patch of skin on the forehead typically forms horizontal wrinkles, while an identical patch of skin between the eyebrows tends to develop vertical wrinkles. This prompt builds on the concept of a `unique identifier' \cite{dream-booth, aging-editing} and can be exemplified by `\textit{image of forehead}''. The UNet reverse process is conditioned on $\tau_\theta(P_{zone})$ to predict the noise level at a given time step $t$, represented as $\epsilon_{zone}$. For this prompt, the regression loss is defined in Equation~(\ref{eq_loss-sign}).

\begin{equation} \label{eq_loss-sign}
    L_{zone} = \mathbb{E}_{t, P_{zone}, \epsilon_t \thicksim \mathcal{N}(0, \mathbf{I})} 
        \Big [ \| \epsilon_t - \epsilon_\theta (z_t, t, \tau_\theta(P_{zone})) \|_2^2 \Big]
\end{equation}

%To ensure consistency between the input and output samples, as required in an unpaired image-to-image translation task, we use a cycle consistency loss \cite{Zhu2017cycle-gan}. In our method, the cycle is computed in the latent space. We use a prompt, $P_{target}$, the same as $P_{full}$ but with the target score, to guide the first step of the cycle consistency loss. However, instead of conditioning the UNet on the current aging score of the zone during the backward process, a different (target) score is used. After this first step, the resulting latent code $\Tilde{z}_0$ undergoes another diffusion process and, to remove this new noise, we use another backward step, conditioning the UNet on the current score of the zone, by using the $\tau_\theta(P_{full})$ latent text prompt. The cycle loss is then computed by comparing the input latent, $z_0$, and the resulting latent from this entire process, $\Bar{z}_0$, with the MSE loss, as defined in Equation~(\ref{eq_loss-cycle}). 

To allow the generation of images from a known source domain with a label $P_{full}$ to a desired target domain with a label $P_{target}$, we introduce a latent cycle consistency loss, inspired by the cycle consistency loss established in CycleGAN\cite{Zhu2017cycle-gan}.

Specifically, starting with the latent representation of an input image, denoted $z_0$, we condition the UNet with $P_{target}$, which represents a prompt specifying a random aging target. Following this initial denoising step, the resulting intermediate latent code, $\tilde{z}0$, is then subjected to a subsequent diffusion process. This process is conditioned to revert to the original source domain $P{full}$, yielding a reconstructed latent output, $\bar{z}_0$. The latent cycle consistency loss is calculated as the Mean Squared Error (MSE) between the initial input latent $z_0$ and its reconstructed counterpart $\bar{z}_0$, as formally defined in Equation~(\ref{eq_loss-cycle}).

\begin{equation} \label{eq_loss-cycle}
    L_{cycle} = \mathbb{E}_{\mathcal{E}(x)} 
        \Big [ \| z_0 - \Bar{z}_0 \|_2^2 \Big]
\end{equation}

For enhanced computational efficiency, we leverage a one-step backward estimation technique. This approximation is given by Equation~(\ref{eq_ddpm}).

\begin{equation} \label{eq_ddpm}
\mathbf{x}_0 \approx \bar{\mathbf{x}}_0 = \frac{\left( \mathbf{x}_t - \sqrt{1 - \bar{\alpha}_t} \, \epsilon_\theta(\mathbf{x}_t) \right)}{\sqrt{\bar{\alpha}_t}}
\end{equation}

%In addition to performing the latent cycle consistency loss, our proposed architecture introduces another distinct feature to speed up the training: the backward processes consist of only a single step, rather than the 1000 steps suggested in the original DDPM model \cite{ddpm}. This single step calculates the difference between the corrupted latent code at a given time step $t$ and the corresponding predicted noise, expressed as $\Tilde{z}_0 = z_T - \Tilde{\epsilon}_t$ and $\Bar{z}_0 = \Tilde{z}_T - \Bar{\epsilon}_t$.

A Convolutional Neural Network (CNN) is employed as a regression network to check if the score prediction is the same as in the $P_{target}$ prompt, after the first backward of the latent cycle consistency block.
This model (that we call ScoreNet) is trained in parallel to fine-tuning the UNet and the resulting score loss, $L_{score}$ is used to optimize both networks in the training architecture. This loss is computed using the MSE Loss between the predicted and normalized target scores from $P_{target}$. Equation~(\ref{eq_loss-score}) represents this loss.

\begin{equation} \label{eq_loss-score}
    L_{score} = \mathbb{E} 
        \Big [ \| s_{target} - \Tilde{s}_{target} \|_2^2 \Big]
\end{equation}

The final loss is described in Equation~(\ref{eq_model-final-loss}) as $L_{LDLA}$ where each loss component is weighted by a hyperparameter $\lambda_i$. By optimizing both the latent diffusion model and the ScoreNet with the score loss, we ensure that the diffusion model weights are also updated based on the accuracy of the predicted scores.

\vspace{-\baselineskip}
\begin{align} \label{eq_model-final-loss}
    L_{LDLA} = \lambda_{full} L_{full} + 
    \lambda_{zone} L_{zone} + \nonumber \\
    \lambda_{cycle} L_{cycle} +
    \lambda_{score} L_{score}
\end{align}

\subsubsection{Inference stage}

The second stage of our method is the inference process. It consists of an image-to-image pipeline \cite{saharia2022} where the input image is translated to the latent space, and the latent is diffused according to a noise strength $\gamma_n = 0.2$. After the diffusion process, the backward step consists of performing $\gamma_{inf} = 40$ denoising steps. We use $\gamma_{inf}$ steps to accelerate the backward process, instead of performing $1000$ steps, as it is typically done in the classic implementation of DDPMs. Additionally, the guidance scale, $\gamma_g = 0.8$, regulates the importance the text prompt has over the backward process in the attention layers of the network. This parameter has a similar effect to `$w$' in the classifier-free guidance model \cite{class-free-guide}, as latent diffusion models use this concept to condition the network on a class, in this case, the text prompt. 

Taking this into account, the denoised latent, $\Bar{z}_0$, passes through the VAE decoder, resulting in a translated crop with the corresponding target score. Since the inference stage is performed for each zone of the individual's face, a few steps are considered to achieve a fully aged face. First, using an off-the-shelf facial landmarks estimator, we compute facial landmarks to identify the location of each zone of the face (e.g. forehead, upper lip, etc.). Once a zone is identified, the corresponding crop is translated to the target score, and this single crop is blended into the full-face image by using feathering. The process is then repeated for every zone, using a uniform aging score or not, as the aging process may vary for different people. In particular, strong facial expressions promote the increase of forehead and crow's feet wrinkles, and smoking cigarettes promotes the increase of upper lip wrinkles \cite{smoking_expressions}. Our approach can capture the heterogeneity of aging by allowing local control.
% In summary, blending each crop at a time creates a smoother face in the end, as some zone crops overlap the others.

In addition to this smooth blending process, a stable diffusion refiner \cite{sd-xl} is used to better blend the crops. This process employs a pre-trained refiner that enhances the quality of the image. In this case, the same image-to-image pipeline is used, but the UNet that denoises the image is the refined one, with a prompt defined as ``\textit{Utra realistic image of a human face}'' with very small strength applied to the reference image. This ensures that the transformed image is not altered, but its quality is increased, resulting in more natural and appealing results.

%% file: sections/4_datasets.tex
\section{Datasets}
\subsection{High Quality Images Dataset}\label{sub-section:dataset}
We assembled a high-quality image dataset as the initial dataset for this work. It contains 6000 high-resolution ($3000 \times 3000$) images of faces, centered and aligned, covering a broad spectrum of ages (18-80), genders, and self-reported ethnicities. Each facial image is divided into crops, representing a specific zone of the face. These crops are scored based on clinical aging atlases tailored to distinct demographic groups \cite{atlas1, atlas2, atlas3, atlas4, atlas5}, rather than being linked to a specific age. The zones considered in this work primarily correspond to areas prone to wrinkle development, including the glabellar region, nasolabial folds, inter-ocular area, upper lip, under-eye region, corners of the lips, and crow's feet.

\subsection{FFHQ Public Image dataset}
\label{sub-section:ffhq-dataset}
We used the FFHQ dataset introduced in \cite{reference-fid-number2} to develop and evaluate our model on a public dataset. The original dataset comprises 70000 images of people displaying various head orientations, facial expressions, genders, and ages. We filtered the dataset to include only people aged 21 and older, with minimal head rotation and facial expressions, resulting in a curated set of 1100 images. We used skin aging atlases \cite{atlas1, atlas2, atlas3, atlas4, atlas5} to annotate the aging scores for each zone. 

%% file: sections/5_results.tex
\section{Results}
\label{sec:results}

\subsection{High Quality Images Dataset}
\begin{figure*}[h]
    \centering
    
    \begin{subfigure}[h]{\textwidth}
    \centering
    \includegraphics[width=0.7\textwidth]{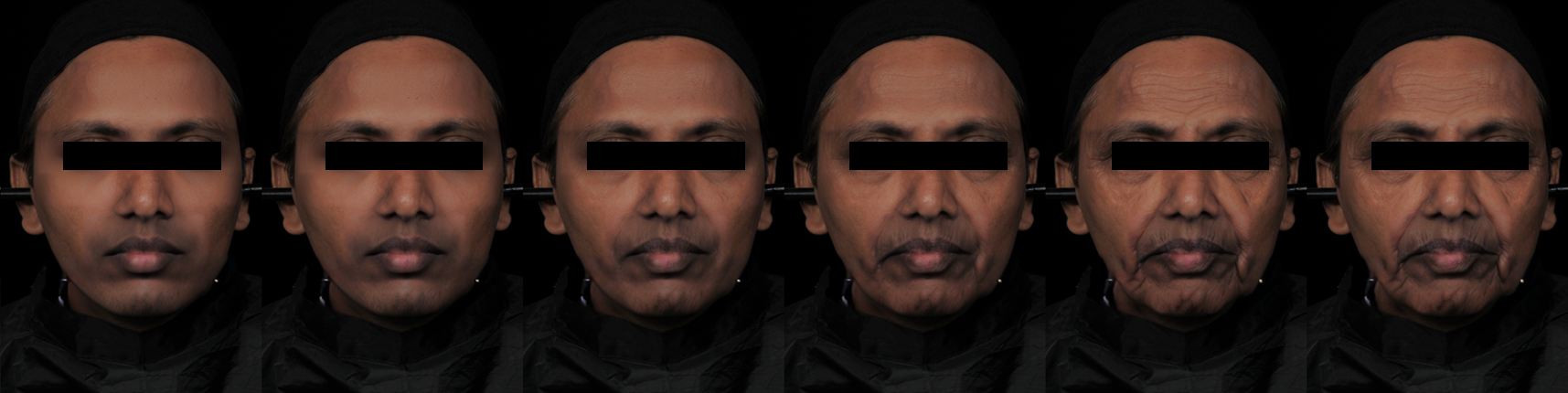}
    \end{subfigure}
    
    \begin{subfigure}[h]{\textwidth}
    \centering
    \includegraphics[width=0.7\textwidth]{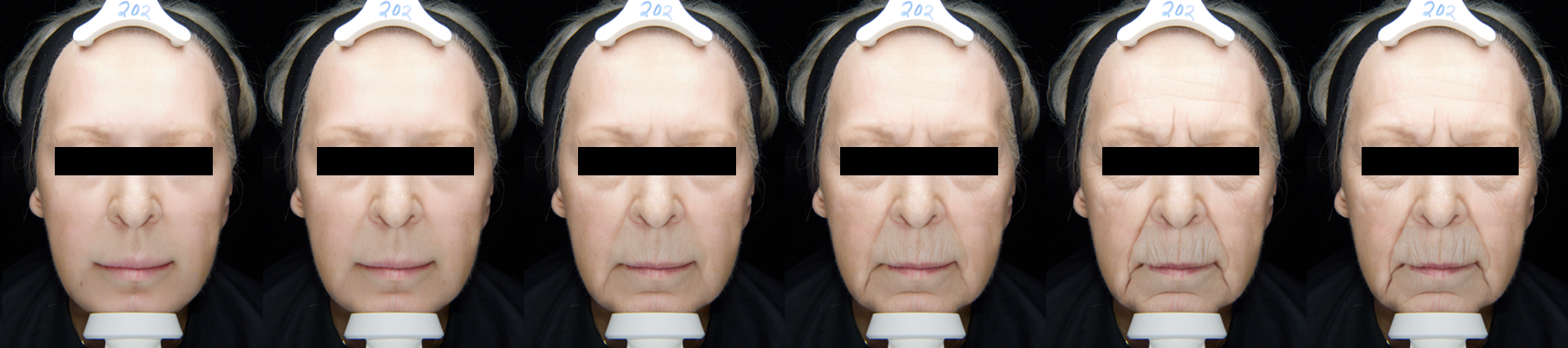}
    \end{subfigure}
    
    \begin{subfigure}[h]{\textwidth}
    \centering
    \includegraphics[width=0.7\textwidth]{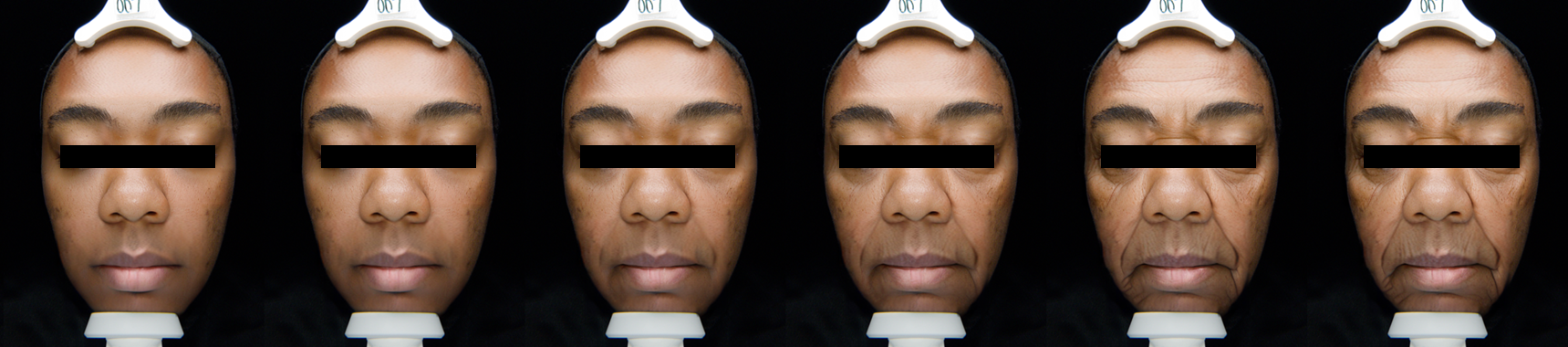}
    \end{subfigure}
    
    \begin{subfigure}[h]{\textwidth}
    \centering
    \includegraphics[width=0.7\textwidth]{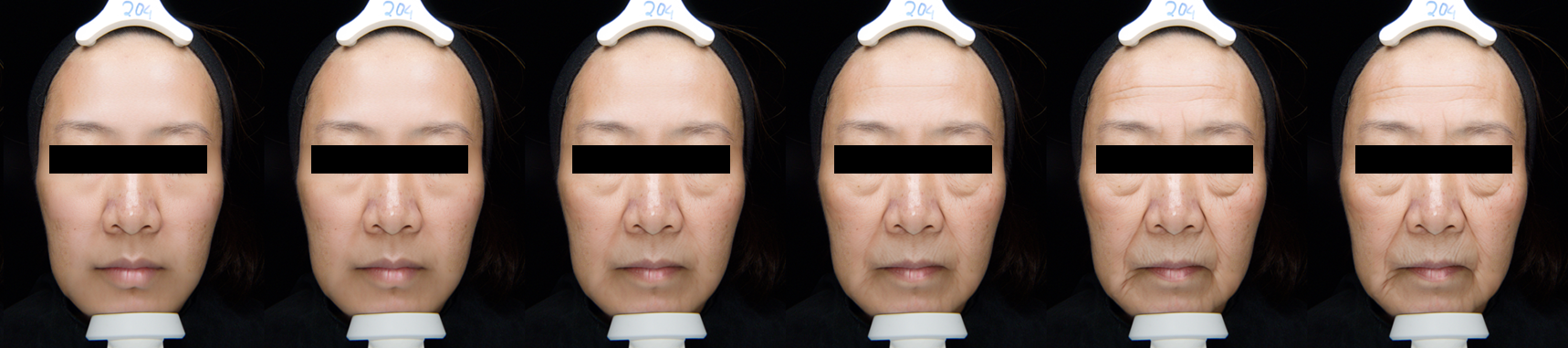}
    \end{subfigure}
    
    \captionsetup{justification=centering, margin=0.7cm}
    \caption{Faces aged with uniform target scores of $5\%, 10\%, 25\%, 40\%, 60\%$ and $85\%$ for different skin types. The faces have been anonymized for licensing reasons.}
    \label{fig:results-1}
\end{figure*}

The images generated with our model exhibit high-quality details and smooth aging transition across the range of skin types and zones. Figure~\ref{fig:results-1} shows our model performs the aging process continuously and smoothly, without artifacts.

\subsubsection{Local control over face wrinkles}
\begin{figure*}[!h]
    \centering
    \includegraphics[width=0.75\linewidth]{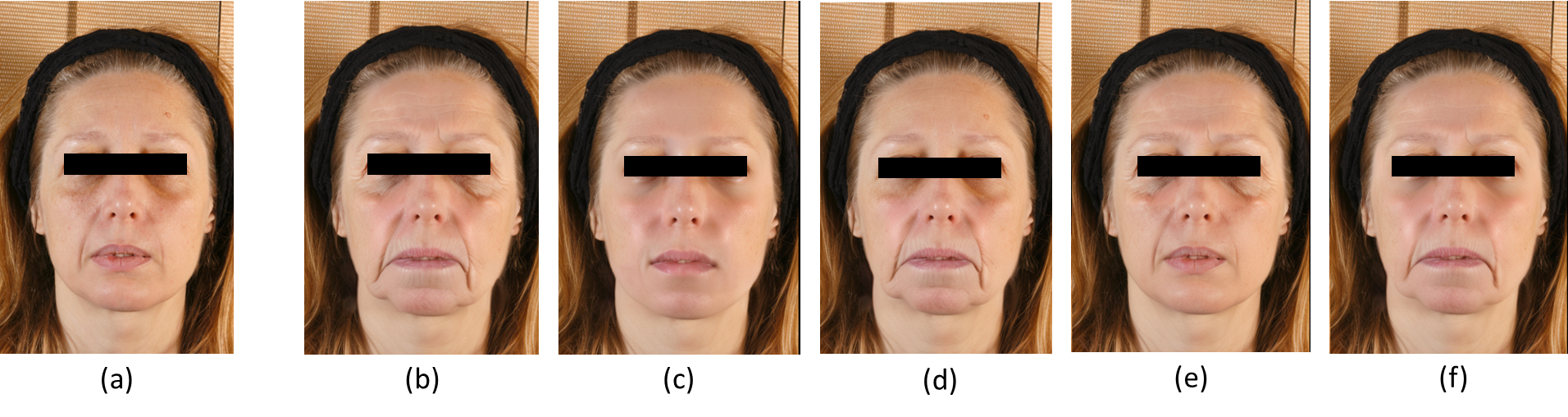}
    \captionsetup{justification=centering, margin=0.7cm}
    \caption{\textbf{(a)} Original image. \textbf{(b)} Entire face aged. \textbf{(c)} Entire face rejuvenated. \textbf{(d)} Bottom face zones aged. \textbf{(e)} Top face zones aged. \textbf{(f)} Some zones have a higher score, others a lower score than (a).}
    \label{fig:playing-zones}
\end{figure*}

To study the degree of control over the aging, Figure~\ref{fig:playing-zones} demonstrates different zones aged with various scores. Compared to classic techniques where the individual ages according to a specific age without considering intrinsic and extrinsic factors, our method offers a more robust aging technique. This approach allows us to define the exact target for each zone, rather than aging the individual according to generic patterns found in the dataset.

\subsubsection{Evaluation metrics}
To succeed in a controlled face aging task, the results must meet three criteria: the image must be realistic, without artifacts; the subject's identity must be preserved; and the face must be aged or rejuvenated to a desired, controlled, target amount. We enforce the first two requirements using the latent cycle consistency loss, an image-to-image pipeline \cite{HF-text2image}, and optimizing the zone loss $L_{zone}$, which ensures the generated crop corresponds accurately to each respective zone. The final requirement is addressed by optimizing the ScoreNet.

When only the first two criteria are respected, we can achieve realism and identity preservation, but fail in aging or rejuvenating. On the contrary, the model could succeed in aging or rejuvenating, but fail in preserving realism or identity.

Based on our experiments, the proposed method effectively preserves the subject's identity without issue. Therefore, we focus on quantitative metrics to evaluate the two remaining criteria. First, we employ the Fr\'{e}chet Inception Distance (FID) \cite{fid} to assess the realism of the generated images. To provide a reference for the score values, we compute this metric between the two halves of the dataset, as the dataset size is significantly smaller than the recommended 50k images \cite{fid, reference-fid-number2}.

\begin{table}[!h] 
\centering
\begin{tabular}{||c c c||} 
     \hline
     \textbf{Model} & \textbf{FID} $\downarrow$ & \textbf{MAE} $\downarrow$ \\ [0.5ex] 
     \hline\hline
     Real images only & $15.65$ & - \\ 
     \hline
     LDLA (Ours) & \textbf{35.71} & $0.15$ \\
     \hline
     AMGAN \cite{amgan} & $45.94$ & \textbf{0.14}\\ [1ex] 
     \hline
\end{tabular}
\captionsetup{justification=centering, margin=0.7cm}
\caption{Comparison between the AMGAN method and the proposed method.}
\label{tab:metrics_results}
\end{table}

We also compute the Mean Absolute Error (MAE), a metric that measures the average absolute difference between the predicted scores and the target scores. We compare our results with the AMGAN model \cite{amgan} and present them in Table~\ref{tab:metrics_results}. Considering the FID metric, our LDLA method excels in realism and produces images with scores closer to the desired ones, with an MSE similar to that of GANs.

\subsubsection{Comparison with GAN baseline}
\begin{figure*}[!h]
    \centering
    \includegraphics[width=0.85\linewidth]{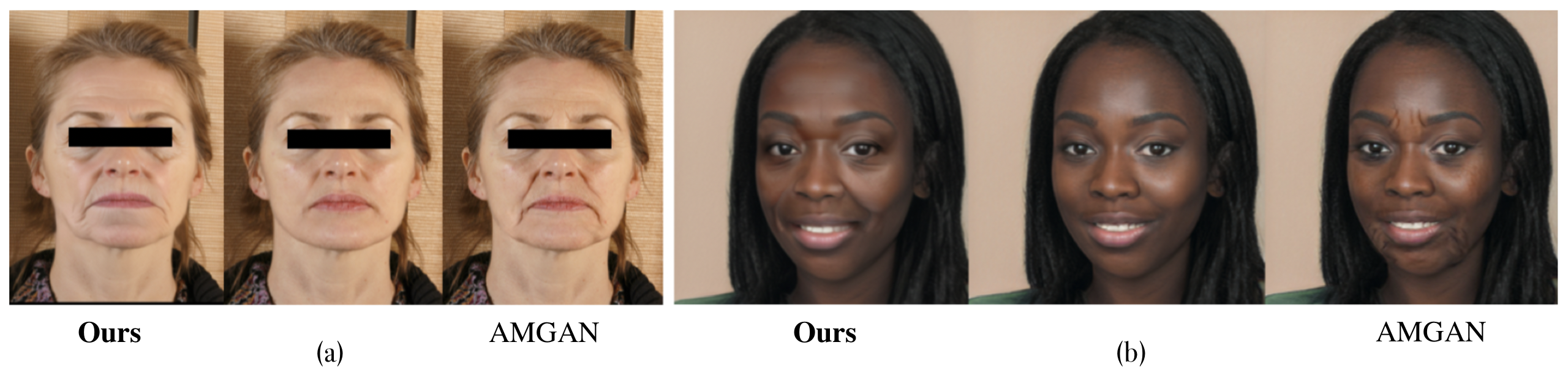}
    \captionsetup{justification=centering, margin=0.7cm}
    \caption{Comparison between our model and AMGAN \cite{amgan} on our High-Quality Image Dataset (a) and FFHQ (b).}
    \label{fig:COMPARE-GANS}
\end{figure*}

Our results are comparable to those of GANs, but our model displays greater generation capabilities when the original individual shows no existing signs of aging (Figure~\ref{fig:COMPARE-GANS}). In that case, the GAN method sometimes fails to produce realistic face wrinkles. This issue is not seen in our LDLA method. One limitation observed in our approach, however, is the occasional presence of blur. This may arise from the resolution of $128 \times 128$ used for crops in our model.

\subsection{Experiments on FFHQ Dataset}
\begin{figure*}[!h]
    \centering
    \includegraphics[width=0.85\linewidth]{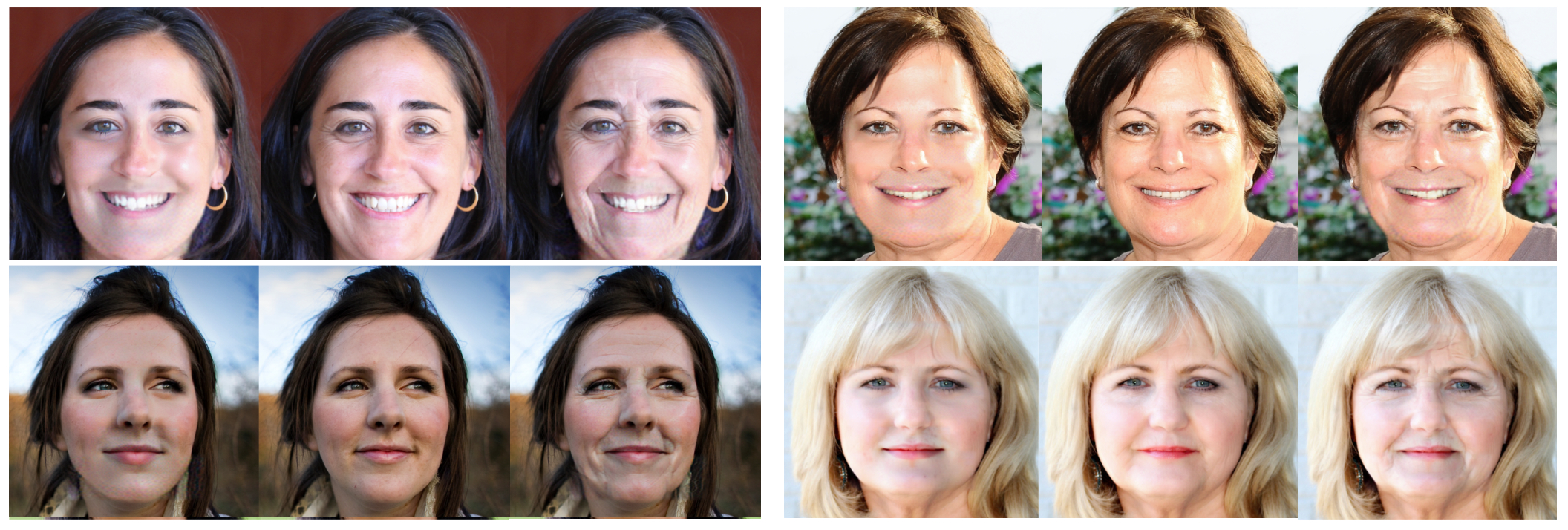}
    \captionsetup{justification=centering, margin=0.7cm}
    \caption{Rejuvenated (left), original (center), and aged (right) using our approach on the  FFHQ dataset.}
    \label{fig:ffhq-generation}
\end{figure*}

To evaluate our model on a more challenging publicly available dataset, we trained it on the subset of FFHQ described in Section~\ref{sub-section:ffhq-dataset}. The results for uniform aging are presented in Figure~\ref{fig:ffhq-generation}. These results show that the model achieves high-quality aging despite the much smaller number of training images, as well as variations in illumination, head orientation, and facial expressions across the dataset. In Figure~\ref{fig:COMPARE-GANS}, we can see that LDLA also outperforms AMGAN on this dataset, yielding more natural, artifact-free images. Note that, due to our crop-by-crop approach, the model still retains the capability to perform non-uniform aging (with different aging scores for each zone), the results presented use uniform aging for ease of visualization. 

\subsection{Ablation Study}
\subsubsection{Effect of losses}
\begin{figure*}[!h]
    \centering
    \includegraphics[width=0.75\linewidth]{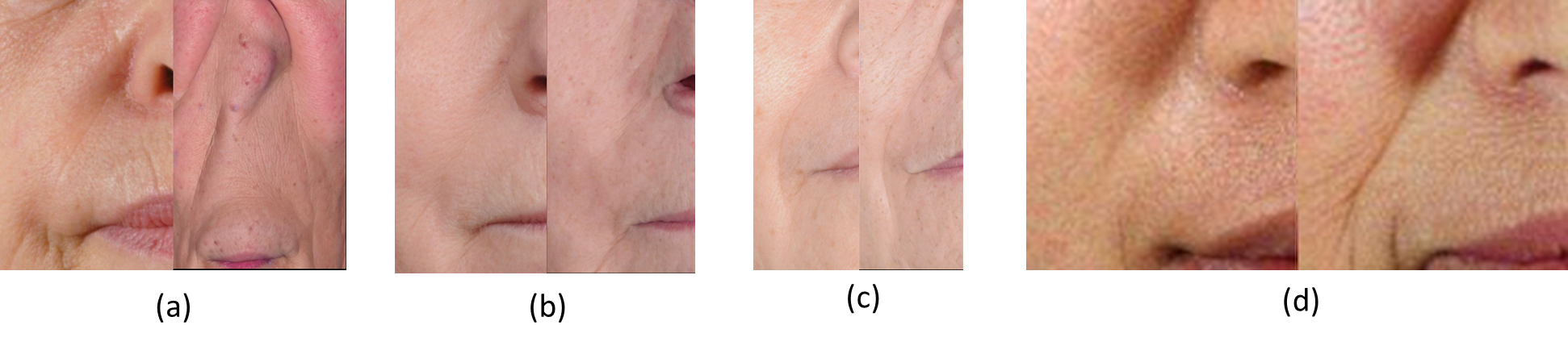}
    \captionsetup{justification=centering, margin=0.7cm}
    \caption{\textbf{(a)} text-to-image pipeline with $L_{full}$. \textbf{(b)} image-to-image pipeline with $L_{full}$. \textbf{(c)} image-to-image pipeline with $L_{full}$ and $L_{zone}$. \textbf{(d)} Full model. In all cases, the left is the reference, and the right is the transformation.}
    \label{fig:ablation2}
\end{figure*}

We observe that, with a simple text-to-image pipeline with $L_{full}$, the model fails in identity preservation, aging, and realism (Figure~\ref{fig:ablation2}a). When a text-to-image pipeline is replaced by an image-to-image pipeline, we notice an improvement in identity preservation. However, the model still fails in aging and realism, as it does not accurately recognize the characteristics of each facial zone (Figure~\ref{fig:ablation2}b). By adding the zone loss, we observe a refinement in realism, but the model fails in identity preservation and controlled aging or rejuvenation of the zone (Figure~\ref{fig:ablation2}c). Finally, our proposed model optimizes the full loss presented in Equation \ref{eq_model-final-loss}. It preserves the identity details of the reference image, achieves realism, and successfully ages or rejuvenates the corresponding zone (Figure~\ref{fig:ablation2}d).

Table~\ref{tab:metrics_results_losses} compares our proposed fine-tuned stable diffusion model with the reference FID and the proposed method without the cycle consistency and the score losses. In this table, we compute the FID using 5 face zones to study the model in different controlled age situations. As expected, the final model outperforms the version without the cycle consistency and score losses.

\begin{table}[!h] 
\centering
\begin{tabular}{||c c||} 
     \hline
     \textbf{Model} & \textbf{FID} $\downarrow$ \\ [0.5ex] 
     \hline\hline
     Real images only & $15.65$ \\ 
     \hline
     LDLA w/o $L_{cycle} + L_{score}$ & $22.85$ \\ 
     \hline
     LDLA (Ours) & \textbf{20.28} \\ [1ex] 
     \hline
\end{tabular}
\captionsetup{justification=centering, margin=0.7cm}
\caption{FID comparison between the model without the cycle consistency and score losses.}
\label{tab:metrics_results_losses}
\end{table}

\subsubsection{Choice of finetuned model}
\begin{figure*}[!h]
    \centering
    \includegraphics[width=0.7\linewidth]{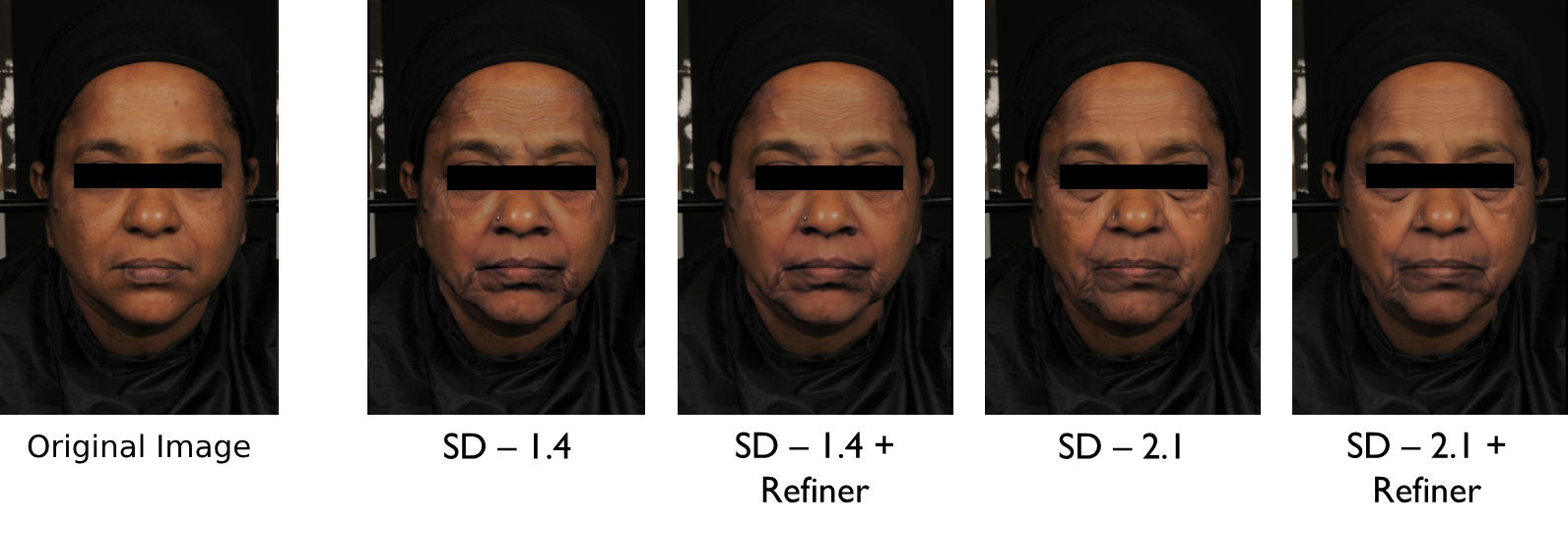}
    \captionsetup{justification=centering, margin=0.7cm}
    \caption{Comparison of fine-tuned models and the use of refiner to enhance generation quality.}
    \label{fig:comparing-models}
\end{figure*}

To justify the choice of fine-tuning the stable diffusion model version 2.1 with a refiner \cite{sd-xl}, instead of version 1.4, Figure~\ref{fig:comparing-models} illustrates the aged version of the individual with and without the refiner and on both models. As expected, thanks to more fine-tuned parameters, the pre-trained version 2.1 leads to better results in terms of skin tone and the natural appearance of aging signs. Additionally, especially in the upper lip and nasolabial regions, it is possible to note small traces of the transformed crop when no refiner is used. In contrast, the refiner allows a smoother transition between the transformed and non-transformed regions.

% \subsubsection{Effect of ethnicity in the prompt}
% The UNet uses a cross-attention mechanism to condition the prompt on the network and guide the denoising process. For this reason, it is important to evaluate the prompt influence over the generated crop. Specifically, it is crucial to observe how ethnicity influences the generation. Figure \ref{fig:ablation1} compares the controlled aging process for two individuals, varying the ethnicity in the prompt. The results demonstrate that the method is robust to changes in the ethnicity specified in the prompt, particularly for Caucasians, the majority in the datasets. However, there are still certain ethnicities that require accurate designation by dermatologists to ensure proper generation. 

% \begin{figure*}[!h]
%     \centering
%     \includegraphics[width=0.60\linewidth]{images/ablation1.png}
%     \captionsetup{justification=centering, margin=0.7cm}
%     \caption{On the left, the original image. On the right, the aged versions with different ethnicities. \textbf{(a)} Individual of `Afro American' ethnicity. \textbf{(b)} Individual of `Caucasian' ethnicity.}
%     \label{fig:ablation1}
% \end{figure*}

\subsection{Limitations of the model}
\label{sec:blurry-limitation}
The full face collected images are of size $1024 \times 1024$. When we perform the blending process, the collected crop must be resized to match the dimensions we trained the model, which is $128 \times 128$. By doing so, it receives a low-quality image that has lost important details. Consequently, the generated image crop might also exhibit low resolution and lack of detail compared to the original collected crop. After the blending process is completed for all crops, the entire face is resized to allow for zooming and detailed observation. However, this final full face might present blurry patches, especially in the regions around the mouth and forehead, where the crops are the largest. 
% Figure \ref{fig:limitation2} shows an example where this effect is particularly noticeable.

A potential solution for this problem could be training the model on crops of larger sizes, such as $256 \times 256$ and $512 \times 512$. However, this approach is more time and memory-consuming.

%% file: sections/6_conclusion.tex
\section{Conclusion}
In this work, we presented a method based on Latent Diffusion Models with several improvements to allow locally controlled face aging using ethnicity-specific skin atlases. With the introduction of custom losses taking into account both general guidance on aging, and task-specific specialization combined with a cycle consistency mechanism, the proposed method has proven to respect the three constraints of face aging: identity preservation, realism, and natural aging.

Our method improves upon previous GAN-based approaches by adding robustness to the absence of pre-defined aging marks on the person’s face as well as by allowing more stable training thanks to the use of Latent Diffusion Models.

Finally, while the approach has been shown to be effective when trained on low resolution crops, improving the input resolution should result in more detailed full-face images. This would come at the expense of additional computational needs and can be seen as a trade-off between image quality and compute requirements.

% Additionally, the method could be extended to use a similar approach to that presented in AMGAN \cite{amgan}, by fine-tuning the UNet to generate an aging mask that is added to the reference image. Combined with a cycle consistency loss, this approach can enhance self-identity preservation.

% In conclusion, our method demonstrates significant potential and can be applied in various situations. Given the control over face wrinkles, it is possible to simulate different scenarios and illustrate the effects of environmental conditions and lifestyle behaviors. The multi-task nature of stable diffusion models allows this approach to be more flexible than models based on GANs and can be extended to a wide range of applications.